\documentclass[journal]{IEEEtran}
\usepackage{amsmath,amsfonts}
\usepackage{algorithm}
\usepackage{array}

\usepackage{url}
\usepackage{hyperref}
\usepackage{xr-hyper}
\usepackage{booktabs}       
\usepackage{nicefrac}       
\usepackage{microtype}      
\usepackage{wrapfig}
\usepackage{multirow}
\usepackage{algpseudocode}
\usepackage{subfigure}
\usepackage{textcomp}
\usepackage{stfloats}
\usepackage{bm}
\usepackage{verbatim}
\usepackage{graphicx}
\usepackage{cite}
\usepackage{multirow}
\usepackage{amssymb}
\usepackage{mathtools}
\usepackage{amsthm}
\usepackage{booktabs} 
\usepackage{enumerate}
\usepackage{float}
\usepackage{soul} 
\usepackage{color, xcolor}
\newtheorem{theorem}{Theorem}
\newtheorem{lemma}{Lemma}

\newtheorem{assumption}{Assumption}

\usepackage{caption}

\makeatletter
\newcommand*{\addFileDependency}[1]{
  \typeout{(#1)}
  \@addtofilelist{#1}
  \IfFileExists{#1}{}{\typeout{No file #1.}}
}
\makeatother

\newcommand*{\myexternaldocument}[1]{%
    \externaldocument{#1}%
    \addFileDependency{#1.tex}%
    \addFileDependency{#1.aux}%
}
\myexternaldocument{supple}
\begin{document}

\title{Learn What You Need \\
in Personalized Federated Learning}
\author{Kexin Lv, Rui Ye, Xiaolin Huang,~\IEEEmembership{Senior Member,~IEEE,}  Jie Yang, Siheng Chen,~\IEEEmembership{Member,~IEEE}
\thanks{K. Lv, X. Huang and J. Yang are with Institute of Image Processing and Pattern Recognition, Shanghai Jiao Tong University, Shanghai 200240, China (e-mail: kelen\_lv@sjtu.edu.cn, xiaolinhuang@sjtu.edu.cn, jieyang@sjtu.edu.cn).}
\thanks{R. Ye is with  the Cooperative Medianet Innovation Center (CMIC), Shanghai Jiao Tong University, Shanghai 200240, China (e-mail: yr991129@sjtu.edu.cn).}
\thanks{S. Chen is with Shanghai Jiao Tong University, Shanghai 200240, China and also with Shanghai AI Laboratory, Shanghai 200232, China (e-mail: sihengc@sjtu.edu.cn).}
\thanks{Correspondence to Jie Yang and Siheng Chen.}}

\markboth{Journal of \LaTeX\ Class Files,~Vol.~14, No.~8, August~2021}%
{Shell \MakeLowercase{\textit{et al.}}: A Sample Article Using IEEEtran.cls for IEEE Journals}


\maketitle

\begin{abstract}
Personalized federated learning aims to address data heterogeneity across local clients in federated learning. However, current methods blindly incorporate either full model parameters or predefined partial parameters in personalized federated learning. They fail to customize the collaboration manner according to each local client's data characteristics, causing unpleasant aggregation results. To address this essential issue, we propose $\textit{Learn2pFed}$, a novel algorithm-unrolling-based personalized federated learning framework, enabling each client to adaptively select which part of its local model parameters should participate in collaborative training. The key novelty of the proposed $\textit{Learn2pFed}$ is to optimize each local model parameter's degree of participant in collaboration as learnable parameters via algorithm unrolling methods. This approach brings two benefits:  1) 
{mathmatically determining the participation degree of local model parameters in the federated collaboration}, and 2) obtaining more stable and improved solutions. Extensive experiments on various tasks, including regression, forecasting, and image classification,  demonstrate that $\textit{Learn2pFed}$ significantly outperforms previous personalized federated learning methods.
\end{abstract}


\section{Introduction}
\begin{figure}[ht]
  \centering
    \includegraphics[scale=0.78]{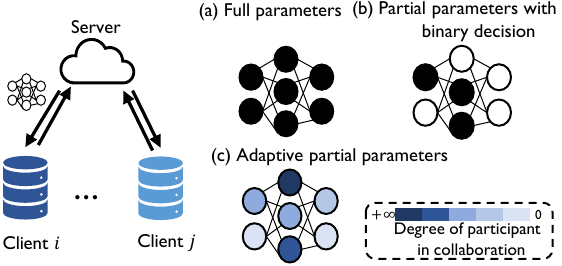}
    \caption{Three federated ways of local model parameters in (personalized) federated learning: sending (a) full parameters; (b) partial parameters with binary decision; (c) adaptive partial parameters. We aim to determine the part and the degree of local model parameters that participate in federated collaboration.}\label{fig:linear}
\end{figure}

\IEEEPARstart{F}{ederated} learning (FL) is an emerging collaboration paradigm that was first introduced in~\cite{fedavg}. 
In the classical FL framework, local clients receive an identical global model from the server and conduct independent local model training on their respective datasets. Subsequently, they send their individual models back to the server for further model aggregation, which is performed there. This iterative process between the server and local clients, as shown in Fig.~\ref{fig:linear}(a), persists until achieving a satisfying global model.
Since only an update to the current global model is uploaded in FL, instead of raw datasets, it can protect data privacy in some degree. Due to this characteristic, it is widely used in finance~\cite{Long2020}, healthcare~\cite{nguyen2022federated}, smart cities~\cite{zheng2022applications}, video surveillance~\cite{10142016} and other fields. 
However, data heterogeneity~\cite{wang2020tackling,kairouz2021advances,abdulrahman2020survey} across local clients creates deviations between local models and the global model so that they cannot reach the consensus, which is also referred to as client drift~\cite{karimireddy2020scaffold}. Besides, there is a lack of solution personalization for practical applications in classical FL~\cite{9743558}.
Hence, personalized federated learning~\cite{9743558,kulkarni2020survey} has been widely explored to train improved local models within the FL framework, instead of relying solely on a global model.

Previous research in personalized federated learning aims to address data heterogeneity across local clients in federated learning by mainly two approaches: 1) personalizing the global model and 2) learning the personalized model. 
The first approach focuses on generalizing the global model and local adaptation. It involves training a single global model that is then applied in downstream tasks, in line with established FL techniques such as FedAvg~\cite{fedavg} and FedProx~\cite{fedprox},  using techniques like finetuning (FT) and knowledge transfer~\cite{kairouz2021advances} across all local model parameters as shown in Fig.~\ref{fig:linear}(a).
While effective, this approach may not fully capture the unique characteristics of individual clients' data.
The second approach, which our method belongs to, aims to provide personalized solutions within the federated learning framework. By modifying the learning process with full model parameters in FL, these model-based personalized FL methods are presented in a variety of ways, e.g., optimization on the well-designed objective functions~\cite{pFedMe, ditto, lp-proj}, meta-learning~\cite{perfedavg}, clustering~\cite{CFL,marfoq2022personalized}, generative networks~\cite{pmlr-v139-shamsian21a}, etc.
Among them, some works~\cite{fedper,fedrep,pmlr-v162-pillutla22a,singhal2021federated,sun2023fedperfix} realize that personalization with full parameters may be unnecessary, and manually divide them into personal parameters and shared parameters, where only the former is updated locally, as illustrated in Fig.~\ref{fig:linear}(b). However, these works do not examine to what degree these chosen partial parameters should be integrated into the federated learning process. The limited variability caused by binary selection hinders the creation of personalized models that could better adapt to local data.

Motivated by this, we aim to learn to determine which part of a local model  should participate in federated learning and further to~\emph{what degree}, as illustrated in Fig.~\ref{fig:linear}(c). To achieve this, our key idea is to consider each parameter’s  degree of participant in collaboration as one learnable variable, and then optimizes those parameters in algorithm unrolling. Following this spirit, we propose a novel algorithm-unrolling-based personalized federated learning framework, $\textit{Learn2pFed}$. 
Specifically, it unrolls the parameters, originally in the iterative algorithm that can indicate the degree of participant in collaboration, into layers of a deep network. Supervised by the sum of training losses collected from all local clients, $\textit{Learn2pFed}$ adaptively learns the characteristics of the local data and select appropriate partial parameters. 

Different from the previous works enabling clients to adaptively share partial model parameters from a whole candidate set, including methods such as parameter prunning~\cite{jiang2022fedmp,sun2021partialfed}, and subnet training~\cite{diao2021heterofl,isik2023sparse}, we adaptively learn local parameters that are fixed in other methods in FL framework, e.g., the aggregated weights, by algorithm unrolling, which makes learning personalized models more challenging.
To sum up, $\textit{Learn2pFed}$ has two distinct advantages: 1) it dynamically determines which parameters of the local model need to collaborate in FL and to what degree, thus adapting to the local data better and improving the performance of personalized FL; and  2) it leverages algorithm unrolling to make hyper-parameters learnable and significantly improves the model capability. 

To evaluate $\textit{Learn2pFed}$, we consider various personalized FL tasks including  regression, forecasting and image classification on different datasets: synthetic polynomial data, power consumption data, Fashion-MNIST~\cite{xiao2017fashion} and CIFAR-10~\cite{krizhevsky2009learning}. $\textit{Learn2pFed}$ outperforms the previous personalized FL methods in the above three tasks.

Our main contributions are three-fold: 
\begin{enumerate}
    \item We introduce adaptive collaboration in personalized federated learning by enabling each client to select which part of its local model parameters should participate in personalized federated learning, addressing data heterogeneity and improving aggregation results. 
    \item We propose a novel algorithm-unrolling-based framework $\textit{Learn2pFed}$ for personalized federated learning to optimize the degree of participant for each model parameter in collaboration, which turns the fixed hyper-parameters in the optimization into learnable parameters in our framework. 
    \item We conduct extensive experiments in various tasks, and show that the performance is competitive with state-of-the-art methods.
\end{enumerate}

This article is organized as the following. Section~\ref{sec:2} reviews the related works. Section~\ref{sec:3} presents several preliminaries about personalized FL. 
Section~\ref{sec:4} introduces our proposed \textit{Learn2pFed} method, including algorithm design, federated implementation, theoretical analysis and extensive discussion. Section~\ref{sec:5} shows the experimental results in three different tasks. And the conclusion goes to Section \ref{sec:con}.

\section{Related Works}\label{sec:2}

\textbf{Personalized Federated Learning:} Personalized federated learning~\cite{9743558} aims to deal with the data heterogeneity and provide personalized solutions. One popular strategy is performing model personalization from the globally shared FL model by finetuning~\cite{kairouz2021advances}, meta-learning (e.g., Per-FedAvg~\cite{perfedavg}, pFedMe~\cite{pFedMe}), model interpolation (e.g., FedProx~\cite{fedprox}, SCAFFOLD~\cite{karimireddy2020scaffold}), transfer learning (e.g., FedMD~\cite{li2019fedmd}, Co-MDA~\cite{10128163}), etc.
Another strategy aims to learn personalized models that involves parameter pruning, sub-network selection and so on. 
Among them, FedPer~\cite{fedper} and FedRep~\cite{fedrep} artificially determine the local model as base layers and personalized layers, and keep the latter private in local training to learn personalized representation.
For different computation and communication capabilities, HeteroFL~\cite{diao2021heterofl}  adaptively allocating local models of different complexity levels based on the global model, and FedPM~\cite{isik2023sparse} learns a binary mask to find the optimal sparse random network within the original one. FedMP~\cite{jiang2022fedmp} performs model pruning, where the server adaptively determines the specific pruning ratio according to the capabilities of local clients. 
And PartialFed-Adaptive~\cite{sun2021partialfed} learns the personalized loading strategy by reparameterization for each client so that the local model is a subset of the global model’s parameters.
In contrast to these works, the proposed \textit{Learn2pFed} is a novel framework closed to meta-learning and aims to better adapt to local data in personalized FL via learning the degree of participant in federated collaboration with algorithm unrolling.  

\textbf{Algorithm Unrolling:} 
Algorithm unrolling~\cite{monga2021algorithm} is a technique that unrolls one specific iterative optimization algorithm, e.g.,  the iterative shrinkage and thresholding algorithm~(ISTA~\cite{beck2009fast}), the alternating direction method of multipliers~(ADMM~\cite{MAL-016}), into stacked layers of a deep network. Then, each forward propagation of the network is equivalent to performing several iterations of the iterative algorithm with fixed parameters. And the backward propagation of the deep model makes the fixed parameters learnable. 
In this way, unrolling enhances both the representation ability of the iterative algorithm and the generalization ability of the generic neural networks, thus reaching an attractive balance.
For these advantages, it has been widely applied in various domains, including the context of sparse coding~\cite{gregor2010learning}, compress sensing~\cite{yang2018admm}, image fusion~\cite{9416456} and signal denoising~\cite{9453145,9414453,8950351}. In our work, we leverage deep unrolling to determine the personal parameters in personalized federated learning, bridging the gap between iterative algorithms and the federated learning framework.

\section{Preliminary}\label{sec:3}
The personalized FL framework consists of one parameter server and $M$ local clients, where the $i$-th  client holds the local data $\mathcal{D}_i=\{X_i, Y_i\}$ with $X_i\in\mathbf{R}^{n_i\times k}, Y_i\in\mathbf{R}^{n_i}$ generated from one of the unknown models. $n_i$ denotes the number of samples in the $i$-th client and  $k$ denotes the feature dimension.
Let $w\in \mathbf{R}^k$ be the global model parameters, and $v_i\in \mathbf{R}^k$ be the $i$-th local model parameters for $i\in[M]$, where we denote the set $\{1, 2, \dots, M\}$ for any integer $M$ as $[M]$.

{Generally, the objective of personalized FL is formed as the local objectives $L_i(v_i; w^*)$ given the optimized global model $w^*$, composed of local empirical loss $F_i(v_i)$ on the local training data in the $i$-th client and the regularized term $\|v_i\!-\!w^*\|^2$ indicating the distance between the global model and local model. 
Mathematically, the optimization of personalized FL is typically formed as below.
\begin{equation}\label{eq:pfl}
\begin{aligned}
    \min_{v_i}\quad  &L_i(v_i;w^*)= F_i(v_i)+\lambda\|v_i-w^*\|^2, \\
\mathrm{s.t.}\quad &w^* = \arg\min_{w} \sum_{i=1}^M p_i L_i(v_i^*;w),
\end{aligned}
\end{equation}
where $\lambda$ and $\{p_i\}$ are two kinds of positive hyper-parameters in personalized FL, and $\{\cdot\}$ denotes the abbreviation of $\{\cdot\}_{i=1}^M$.
Specifically, $\lambda$ regularizes the similarity between the global model and local models, with larger values of $\lambda$ indicating stronger similarity. 
When $\lambda\!\to\!\infty$, personalized FL degrades to the general FL; when $\lambda\!=\!0$, personalized FL degrades to the local independent learning.} 

While (\ref{eq:pfl}) provides the mathematical form commonly used in personalized federated learning methods, it has a limitation arisen from treating the entire local parameter model as a single entity, thus overlooking the unique characteristics of local data.
This limitation may hamper the ability to adapt the model to individual data distributions and can result in worse performance in personalized federated learning.
Therefore, it becomes crucial to address this limitation and develop a solution  by learning the specific parameters of local models in collaboration.

In this regard, we propose $\textit{Learn2pFed}$, a novel framework that entails redesigning the formulation of  (\ref{eq:pfl}). We will delve into the details of the $\textit{Learn2pFed}$ framework in the next section.

\section{\textit{Learn2pFed}: Unrolling-based Personalized FL Framework}\label{sec:4}
To determine which specific parameters of the local models should participate in the federated learning, this section introduces $\textit{Learn2pFed}$, a novel deep unrolling framework for personalized federated learning from both aspects of mathematical optimization and federated implementation. We further provide its convergence analysis, and discuss its characteristics including privacy, computation burden, and communication cost.

\subsection{Overall Optimization}
{Based on the original optimization problem (\ref{eq:pfl}), we introduce another crucial component, ${\Lambda_i}$, alongside the aggregation weight variable $p_i$. This addition allows us to achieve personalized regularization for each model parameter, further enhancing the adaptive federated aggregation.}

\textbf{Regularized variable $\Lambda_i$:} Instead of using a scalar $\lambda$ in (\ref{eq:pfl}) to regularize all model parameters, we introduce a personalized diagonal matrix $\Lambda_i \in \mathbf{R}^{k\times k}$ for the $i$-th client for element-wise regularization. Each element $(\Lambda_i)_{jj}$ is a positive value, indicating the degree of each model parameter that participates in the federated collaboration. This matrix enables customized regularization for different parameters within each client's local model. Such fine-grained personalized regularization allows for adaptive control of the degree of participant in collaboration, improving model performance by tailoring the regularization to the specific characteristics of each client's data.


Subsequently, the overall optimization of $\textit{Learn2pFed}$ is formulated as a bi-level optimization problem, which involves the learning objective $\mathcal{P}_f(\{v_i\},w)$ and the constraint problem $\mathcal{P}_b(\{\Lambda_i,p_i\})$:
\begin{equation}\label{eq:loss}
\begin{aligned}
    \min\limits_{\{v_i\},w}&\mathcal{P}_f(\{v_i\}\!,\!w)   \!=\!\frac{1}{M}\sum_{i=1}^M p_i\left(F_i(v_i)\!+\!(v_i\!-\!w)\!^\top\!\Lambda_i(v_i\!-\!w)\right)\\
     \mathrm{s.t.}\quad& \{\Lambda_i,p_i\}=\arg\min\limits_{\{\Lambda_i,p_i\}}~ \mathcal{P}_b(\{\Lambda_i,p_i\})\!=\!\sum\nolimits_{i=1}^M F_i(v_i^\star),
\end{aligned}
\end{equation}
{where $v_i^\star$ is the output of $\mathcal{P}_f(\{v_i\},w)$} and $F_i(v_i^\star)$  denotes the  local training loss in the $i$-th client based on the specific tasks, such as Mean-Squared-Error (MSE) loss for regression or Cross-Entropy (CE) loss for classification. 
Intuitively, (\ref{eq:loss}) aims to output the learned local model $\{v_i\}$ for personalized FL, while learning the adaptive collaboration pattern via  learnable parameters $\{\Lambda_i,p_i\}$ with the supervised information in the form of the sum of local training losses.
Unlike (\ref{eq:pfl}), (\ref{eq:loss}) also includes the learning of $\{\Lambda_i,p_i\}$, thus it can adaptively determine the specific part of local model parameters involved in the collaboration, allowing for a more flexible and effective personalized federated learning process.

To address the optimization problem presented in (\ref{eq:loss}), we leverage algorithm unrolling. Specifically, our approach involves solving the objective of (\ref{eq:loss}) using a single optimization algorithm, as discussed in Section~\ref{sec:opt}. Subsequently, we unroll this algorithm into layers and train a deep network, as explained in Section~\ref{sec:unroll}. 
\vspace{-2mm}
\subsection{Optimization Algorithm}\label{sec:opt}
This sub-section aims to solve the learning objective of (\ref{eq:loss}) with fixed parameters $\{\Lambda_i,p_i\}$.
Since the global model and local models are coupled in $\mathcal{P}_f(\{v_i\},w)$ in (\ref{eq:loss}), the alternating direction method of multipliers (ADMM~\cite{MAL-016}) is a way to split the variables into local sides and the global side. 
Specifically, we introduce the auxiliary variable $\{z_i\}$ indicating the consensus constraint in the local. It brings two benefits: 1) it decouples the global and local model so that solving the local variables can be carried out in parallel in each client; 2) it allows for a more flexible expression of constraints making the problem easier to solve.
Then, $\mathcal{P}_f(\{v_i\},w)$ in (\ref{eq:loss}) is reformulated as below. 
\begin{equation}\label{eq:reglob}
\begin{aligned}
        \min_{\left\{z_{i}\right\}, \left\{v_{i}\right\}, w}\quad &\mathcal{P}_f({z_i},{v_i},w)=\frac{1}{M}\sum\nolimits_{i=1}^M {p_i}\left(F_i(v_i)+z_i^\top\Lambda_i z_i\right) \\
        \mathrm{s.t.}\quad &z_i=v_i-w.    
\end{aligned}
\end{equation}
For faster convergence, we also provide its augmented Lagrangian as 
\begin{equation}\label{eq:rho}
\begin{aligned}
        &\mathcal{L}_{p_i,\rho_i,\Lambda_i}(\left\{v_i\right\},\left\{z_i\right\}, w;\{\alpha_i\})\\
        =&\frac{1}{M}\sum\nolimits_{i=1}^M {p_i}\left(F_i(v_i)\!+\!z_i^\top\Lambda_i z_i\!+\!\frac{\rho_i}{2}\left\|z_i-v_i+w+ \alpha_i\right\|^2\right),
\end{aligned}
\end{equation}
where $\{\alpha_i\}$ are Lagrangian multipliers in the local and $\{\rho_i\}$ are positive hyper-parameters.
That is, taking the regression problem as example where $F_i(v_i)=\|X_iv_i-Y_i\|^2$, the ADMM alternatively optimizes $\left\{v_i\right\},\left\{z_i\right\}, w,\{\alpha_i\}$ by solving the following sub-problems in the $\ell$-th iteration. 
 \begin{align}
    &\begin{array}{rl}
         v_i^{\ell}:=\arg\min\limits_{v_i}& \|X_iv_i\!-\!Y_i\|^2\!+\!\frac{\rho_i}{2}\left\|z_i^{\ell\!-\!1}\!+\!w^{\ell\!-\!1}\!+\!\alpha_i^{\ell\!-\!1}\!-\!v_i\right\|^2,
         \label{eq:admm1}  
    \end{array}\\
    &\begin{array}{rl}
         z_i^\ell:=\arg\min\limits_{z_i}& z_i^\top\Lambda_i z_i+\frac{\rho_i}{2}\left\|z_i-v_i^{\ell}+w^{\ell-1}+\alpha_i^{\ell-1}\right\|^2,
         \label{eq:admm2}  \\
    \end{array}\\
    &\begin{array}{rl}
         w^\ell:=\arg\min\limits_{w}& \sum\limits_{{i}}^M \frac{p_i \rho_i}{2}\left\|z_i^{\ell}-v_i^{\ell}+w+\alpha_i^{\ell-1}\right\|^2,
         \label{eq:admm3} 
    \end{array}\\
    &\begin{array}{rl}
         \alpha_i^\ell :=& \alpha_i^{\ell-1}+\rho_i\left(z_i^\ell-v_i^\ell+w^\ell\right). \label{eq:admm4}
    \end{array}
  \end{align}
Since it follows the standard ADMM, its convergence is guaranteed by \cite{hong2016convergence,hong2017linear}.
After performing multiple iterations, e.g., $L$ iterations, till convergence as described above, we obtain the local model $\{v_i^L\}$.

However, determining  $\{\Lambda_i,p_i,\rho_i\}$ plays a critical role in (\ref{eq:admm1})-(\ref{eq:admm4}) since each has a distinct impact on the performance and convergence behavior. For example, the elements of $\Lambda_i$ control the similarity between local and global models for specific features. Tuning these elements influences the models' behavior in capturing global patterns. Similarly, $\rho_i$ affects the convexity and underfitting of local models. Balancing $\rho_i$ is crucial to avoid overfitting or excessive similarity. But selecting suitable $\{\Lambda_i,p_i,\rho_i\}$ is challenging due to their interplay and sensitivity. Manual tuning is time-consuming and prone to biases. Moreover, directly optimizing them in the original problem is not feasible for the trivial solution. Therefore, how to determine $\{\Lambda_i,p_i,\rho_i\}$ is a big challenge, and we provide our method making them learnable in the next section.

\subsection{Algorithm Unrolling}\label{sec:unroll}
\begin{figure*}[ht]
\begin{center}
\includegraphics[scale=0.63]{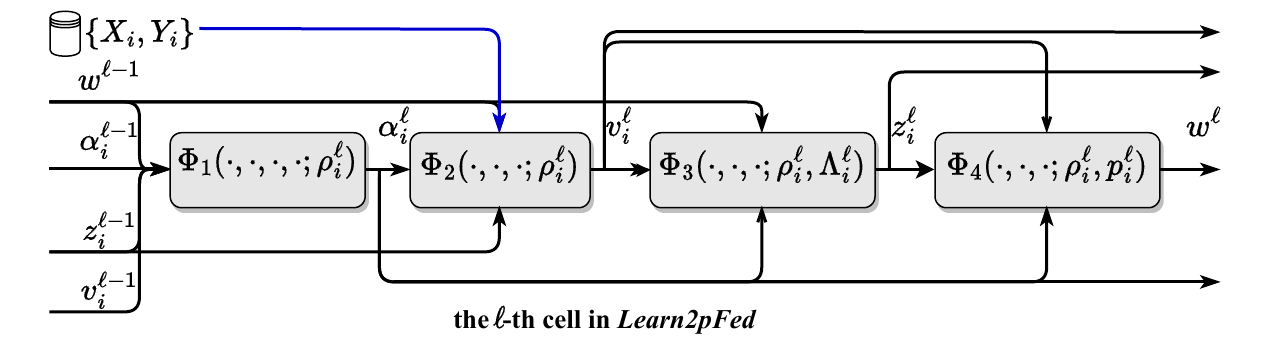}
\caption{Illustration of the $\ell$-th cell in $\textit{Learn2pFed}$. It unrolls (\ref{eq:layer1})-(\ref{eq:layer4}) into one four-layer cell of the deep network. Black lines indicate the flow of intermediate variables, e.g., $\{w^\ell,\alpha^\ell, z_i^\ell, v_i^\ell\}$ in the $\ell$-th cell. Blue line indicates the local data flow, which however, will not be shared across clients.}
\label{fig:1}
\end{center}
\end{figure*}

We introduce the proposed personalized FL framework $\textit{Learn2pFed}$ based on algorithm unrolling to adaptively determine the learnable parameters in the above section. 
The key idea is to view the parameters $\Theta^\ell = \{\Lambda_i^\ell,p_i^\ell,\rho_i^\ell\}$ in (\ref{eq:admm1})-(\ref{eq:admm4}) as trainable parameters in a deep network with the input and parameters as $\Phi(\{X_i,Y_i\}; \{\Theta^\ell\}_{\ell=1}^L)$, where local data $\{X_i,Y_i\}$ are privately stored in local clients.
Specifically, we solve the optimization in (\ref{eq:admm1})-(\ref{eq:admm4}) iteratively, and model one of iterations as a four-layer cell in $\Phi(\{X_i,Y_i\}; \{\Theta^\ell\}_{\ell=1}^L)$, as illustrated in Fig.~\ref{fig:1}.  
Mathematically, we provide the formulations of the intermediate outputs in the $\ell$-th cell as follows.
\begin{align}
    \alpha_i^\ell\leftarrow &~{\Phi_1(\alpha_i^{\ell-1},v_i^{\ell-1},z_i^{\ell-1},w^{\ell-1};\rho_i^\ell)}\nonumber\\ =&\alpha_i^{\ell-1}+\rho_{i}^\ell(z_i^{\ell-1}-v_i^{\ell-1}+w^{\ell-1}),\label{eq:layer1}\\
    v_i^{\ell}\leftarrow &~{\Phi_2(\alpha_i^\ell,z_i^{\ell-1}, w^{\ell-1};\rho_i^\ell)}\nonumber\\
    =&(X_i^\top X_i+\rho_i^{\ell} \mathcal{I}_k)^{-1}(\rho_i^{\ell}(w^{\ell-1}+z_i^{\ell-1}+\alpha_i^{\ell})+X_i^\top Y_i),\label{eq:layer2}\\
    z_i^\ell\leftarrow &~{\Phi_3(\alpha_i^\ell,v_i^{\ell},w^{\ell-1};\rho_i^\ell, \Lambda_i^{\ell})}\nonumber\\
    =&\rho_i^{\ell}(\text{ReLU}(\Lambda_i^{\ell})+\rho_i^{\ell}\mathcal{I}_{k})^{-1}\left(v_i^{\ell}-w^{\ell-1}-\alpha_i^{\ell}\right),\label{eq:layer3}\\
    w^\ell\leftarrow &~{\Phi_4(\alpha_i^\ell,v_i^{\ell},z_i^\ell;\rho_i^\ell, p_i^{\ell})}=\frac{\sum_i p_i^{\ell} \rho_i^{\ell}\left(v_i^{\ell}-z_i^{\ell}-\alpha_i^{\ell}\right)}{\sum_i p_i^{\ell}\rho_i^{\ell}},\label{eq:layer4}
\vspace{-2mm}
\end{align}
where $I_k$ means the identity matrix with the dimension $k$, and the parameters $\Theta^\ell=\{\Lambda_i^{\ell},p_i^{\ell},\rho_i^{\ell}\}$ are learnable. This is the main difference from the iterative algorithm in (\ref{eq:admm1})-(\ref{eq:admm4}). In addition, we build up the ReLU~\cite{lecun2015deep} module to guarantee  the diagonal element of $\{\Lambda_i^{\ell}\}$ is positive, which is given manually in the previous.
In this way, the proposed $\textit{Learn2pFed}$   concatenates multiple four-layer modules as described above into the deep network.
It is worth noting that the update sequence of the ADMM has little impact on its convergence, hence the decision to design the layers is for easier federated implementation.

In the training stage of $\textit{Learn2pFed}$, we consider the following optimization problem:
 \vspace{-1mm}
\begin{equation}\label{eq:back}
    \min_{\{\Lambda_i,p_i, \rho_i\}}\quad \mathcal{P}_b(\{\Lambda_i,p_i, \rho_i\})=\sum\nolimits_{i=1}^M F_i(v_i^L),
\vspace{-2mm}
\end{equation}
where $F_i(v_i^L)$ is the local training loss based on the output of the final layer. 
In contrast to $\mathcal{P}_b(\{\Lambda_i,p_i\})$ in (\ref{eq:loss}), $\{\rho_i\}$  introduced by the ADMM is also treated as the target variable in (\ref{eq:back}).
Then, the parameters $\{\Lambda_i,p_i,\rho_i\}$ are updated iteratively through the standard gradient descent.

In conclusion, the proposed $\textit{Learn2pFed}$ framework performs the iterative algorithm in forward propagation, and trains the learnable parameters in the deep network supervised by the sum of local training losses, which carries high-level information from other clients. 
$\textit{Learn2pFed}$ enjoys the following benefits:
1) it adaptively learns $\{\Lambda_i,p_i, \rho_i\}$ during the training process which could not be learned from optimization aspect otherwise, enabling it to determine the degree of participation of each local model's parameters in the collaboration. This adaptive learning capability allows the framework to dynamically adjust the collaboration strategy based on the specific characteristics of the data and the optimization problem at hand.
2) the integration of deep neural networks in $\textit{Learn2pFed}$ provides a powerful modeling capability. By leveraging the expressive power of deep networks, the framework can capture complex  patterns in each local data, leading to improved performance as demonstrated in Section~\ref{sec:5}.
\subsection{Federated Implementation}\label{sec:fed_imp}
We provide a detailed federated implementation of $\textit{Learn2pFed}$. We initialize local learnable parameters $\{\Lambda_i^{\ell-1}, \rho_i^{\ell-1}\}$, the local model, and its intermediate variables $\{v_i^{\ell-1}, z_i^{\ell-1},\alpha_i^{\ell-1}\}$ on the client sides, where $\ell=1$. Additionally, we initialize global learnable parameters $\{p_i^{\ell-1},\gamma_i^{\ell-1}\}$ and the global model $w^{\ell-1}$, with $\gamma_i^{\ell-1}$ serving as a copy of $\{\rho_i^{\ell-1}\}$ on the server side. We then introduce the implementation on both client and server sides.

1) \textbf{\textit{Client-Side Computation and Communication:}}
In the client sides, $\textit{Learn2pFed}$ updates the intermediate variables $\{\alpha_i^\ell, v_i^\ell, z_i^\ell\}$ by (\ref{eq:layer1}), (\ref{eq:layer2}), (\ref{eq:layer3}), respectively, in the $\ell$-th cell of the deep network based on the learnable parameters $\{\Lambda_i^{\ell}, \rho_i^{\ell}\}$. 
Note that when updating $\{v_i^\ell\}$, 
since $F_i(v_i)$ can be convex or non-convex, we need to discuss the solution separately, and take the two tasks that we will face in the experiments for example.
In regression tasks, we perform (\ref{eq:layer2}) directly.
However, in classification tasks, $F_i(v_i)$ is non-linear. Then we reformulate the update of $\{v_i^\ell\}$ in (\ref{eq:layer2}) using the gradient descent as follows.
\begin{equation}\label{eq:v_2}
    v_i^\ell \leftarrow v_i^{\ell-1} - lr*{\partial h_i(v_i)}/{\partial v_i},
       \vspace{-1mm}
\end{equation}
where we denote $$h_i(v_i)=F_i(v_i)+\frac{\rho_i^\ell}{2}\left\|z_i^{\ell-1}+w^{\ell-1}+\alpha_i^{\ell}-v_i\right\|^2$$ based on (\ref{eq:admm1}).
We find that the approximation accuracy of the solution in this layer does not affect the convergence of the network much in practice, so the learning rate $lr$ can be artificially set. 

As for the communication, 
each local client sends the vector $v_i^{\ell} - z_i^{\ell} - \alpha_i^{\ell}$ and the local training loss $F_i(v_i^L)$  to the server in each cell $\ell \in [L]$ and the final cell $L$ of the network, respectively.
Additionally, each client receives the global model $w^\ell$ and the sum of local training losses across clients broadcasted by the server in each cell $\ell \in [L]$ and the final cell $L$ of the network, respectively.
Finally, each client leverages the sum of losses to independently update their learnable parameters $\{\Lambda_i^\ell, \rho_i^\ell\}$ using the gradient descent method in the final cell $L$ of the network in the client sides.

2) \textbf{\textit{Server-Side Computation and Communication:}} In the server side, $\textit{Learn2pFed}$ updates the intermediate variable $w^\ell$ by (\ref{eq:admm4}) in the $\ell$-th cell of the deep network based on the learnable parameters $\{p_i^\ell, \rho_i^\ell\}$. Since $\{\rho_i^\ell\}$ appears in both sides, we copy it as $\gamma_i^\ell$ and update the only in the server side.
In terms of communication, the server broadcasts the updated global model $w^\ell$ and the sum of local training losses back to all the clients in each cell $\ell\in[L]$ ann the final cell $L$ of the network. At the same time, the learnable parameters $\{p_i^{\ell},\gamma_i^\ell\}$ are updated based on the sum of local training losses in the server side using the gradient descent method in the final cell $L$ of the network.

Then, the above computations and communications are repeated untill $\textit{Learn2pFed}$ converges. To sum up, we summarize the overall algorithm as below.
\begin{algorithm}
\renewcommand{\algorithmicrequire}{\textbf{Input:}}
\renewcommand{\algorithmicensure}{\textbf{Output:}}
\caption{$\textit{Learn2pFed}$: layer-wise training.}\label{alg:1}
\begin{algorithmic}[1]
        \Require The number of local clients $M$; the number of ADMM iterations $L$; the maximum epoch $E$ for training.
        \Ensure Personalized local model $\{v_i^L\}_{i\in [M]}$. 
        \State  Initialize personalized models $\{v_i^0, z_i^0, \alpha_i^0\}$ randomly, and global model $w^0$. Initialize learnable parameters $\{\rho_i^0,  \gamma_i^0, \Lambda_i^0, p_i^0\}$. \Comment{\textbf{Initialization}}
        \For{$e=1$ to $E$}
            \For{$\ell=1$ to $L$}
                \For{$i=1$ to $M$ (parallel)} 
                \Statex \Comment{\textbf{Client-Side Computation}}
                \State Update $\alpha_i^\ell$ via (\ref{eq:layer1}). 
                \State Update $v_i^\ell$ via (\ref{eq:layer2}).
                \State Update $z_i^\ell$ via (\ref{eq:layer3}).
                \State Send vector $vec = v_i^\ell-z_i^\ell-\alpha_i^\ell$ to the server.
                \EndFor
            \State Update $w^\ell$ via (\ref{eq:layer4}) and broadcast it to the local. 
            \Statex \Comment{\textbf{Server-Side Computation}}
            \EndFor
            \State The server collects each local training loss  $\mathcal{L}_i(X_i{v}_i^L,Y_i)$ in (\ref{eq:back}) and updates the global learnable parameters $\{p_i^L,\gamma_i^L\}$.  Then, the server broadcasts the sum of losses back to the clients. 
            \Statex \Comment{\textbf{Global Learnable Parameters Update}}
            \State Each local client receives the losses from the server, and update the learnable parameters both in the client sides. \Comment{\textbf{Local Learnable Parameters Update}}
        \EndFor\\
    \Return $\{v_i^L\}$ after $E$ epochs.
\end{algorithmic}
\end{algorithm}

\subsection{Theoretical analysis}
This sub-section provides theoretical convergence analysis of \textit{Learn2pFed}, including the required assumptions, lemmas and the derived theorem. 
Existing theoretical analyses of the convergence properties of federated learning (FL) algorithms often highlight the dependency of convergence bounds on hyper-parameters. In contrast, our approach features learnable parameters, making it challenging to provide a precise bound. As a result, our convergence analysis is divided into two components: one focuses on the convergence of forward optimization under given parameters, and the other delves into the convergence of stochastic optimization based on SGD~\cite{robbins1951stochastic}. Our basic idea is that the forward propagation of \textit{Learn2pFed} converges to a stationary point given the learnable parameters, and simultaneously, the backward propagation of \textit{Learn2pFed} probabilistically converges to a local minimum. Consequently, the entire network exhibits convergence.

First, we present the relationship between the primal variable $\{v_i\}$ and the dual variable $\{\alpha_i\}$ in the forward propagation of \textit{Learn2pFed} in the following lemma, and its detailed proof goes to  the supplemental materials.
\begin{lemma}\label{le:1}
There exists the positive constant $L_i$ for $\forall i\in [M]$, such that
    $$\|\alpha_i^{\ell}-\alpha_i^{\ell-1}\|\le L_i \|v_i^{\ell}-v_i^{\ell-1}\|. $$
\end{lemma}
Then, we present the following assumptions, which serve as the foundation for our main theorems to follow.
\begin{assumption}\label{ass:0}
Suppose the hyper-parameters $\Theta$ in the forward propagation satisfy the following conditions:
\begin{enumerate}
  \item   The hyper-parameters $\rho_i$ for $\forall i\in[M]$ are large enough so that the $\{v_i\}$-subproblem is strongly convex with modulus $\gamma_i(\rho_i)$, which is a monotonic increasing function of $\rho_i$.
  \item For any $i\in[M]$, 
  the positive constant $L_i$ satisfies $L_i\le\frac{1}{2\overline{p_i}}$, where $\overline{p_i}$ is the maximum value of the finite series $\{p_i^{e}\}$ for $e\in [E]$, so that
  $\gamma_i(\rho_i)\ge 2p_i\rho_iL_i$ by setting $\gamma_i(\cdot)$ as the linear function.
\end{enumerate}
\end{assumption}
Under Assumption~\ref{ass:0}, we can obtain the local convergence of the backward propagation of $Learn2pFed$, which is fulfilled by SGD, with high probability to a minima of the objective function (cf. Theorem 3 and 5 in \cite{fehrman2020convergence}). 
Then, we focus on the convergence of the forward propagation of $Learn2pFed$  under the fixed $\Theta^e$~(simplified as $\Theta$) in the following.
\begin{theorem}\label{theo:1}(Convergence of forward propagation.)
    Suppose Assumption \ref{ass:0} is satisfied, $\{v_i^\ell\}$, $\{z_i^\ell$\}, $w^\ell$ and $\{\alpha_i^\ell\}$ are denoted as the updates obtained at the $\ell$-th forward iteration of \textit{Learn2pFed}.
     We have the following:
\begin{equation*}
\begin{aligned}
	&\mathcal{L}_{\Theta}\left(\left\{v_i^\ell\right\}\!,\!\left\{z_i^\ell\right\}\!,\! w^\ell\!;\!\{\alpha_i^\ell\}\right)\!-\!\mathcal{L}_{\Theta}\left(\left\{v_i^{\ell\!-\!1}\right\}\!,\!\left\{z_i^{\ell\!-\!1}\right\}\!, \!w^{\ell\!-\!1}\!;\!\{\alpha_i^{\ell\!-\!1}\}\right)\\
	\le&\frac{1}{M}\sum_{i=1}^M \left(\left(p_i\rho_i L_i- \frac{\gamma_i(\rho_i)}{2}\right)\|v_i^{\ell}-v_i^{\ell-1}\|^2\right. \\
 &\left. -\frac{\mu}{2}\sum_{i=1}^M p_i\|z_i-z_i\|^2-\frac{\rho}{2}\|w^{\ell}-w^{\ell-1}\|^2\right)\le 0.
\end{aligned}
\end{equation*}
Since the Lagrangian function value is decreasing and lower-bounded, \textit{Learn2pFed} converges as $\ell\to\infty$.
\end{theorem}

The detailed proof goes to the supplemental materials.
And the theorem indicates that the forward propagation of \textit{Learn2pFed} follows the standard ADMM iterates and has a convergent  subsequence. Besides, every limit point is a stationary point for the non-convex problem according to \cite{hong2016convergence},  which does not impose any assumptions on the iterates. To this end, we provide the convergence of the proposed algorithm.

\subsection{Further Discussions}
\textit{1) Privacy:}
Though FL framework avoids local data being exposed, the full model parameters may still leak the data privacy by various attack methods~\cite{fredrikson2015model, shokri2017membership}. However, these attacks primarily rely on the feature reconstruction of individual samples. In \textit{Learn2pFed}, we adhere to the standard vertical Federated Learning (FL) framework, where the gradient is typically computed as the average over a subset of the training data. Moreover, we enhance security by transmitting the estimation of model parameters through a mixed combination of multiple local variables, thereby increasing the difficulty of these attacks.

\textit{2) Computation and Storage Burden:} Our approach introduces auxiliary variables and dual variables in the forward propagation, thereby increasing computational and storage costs. While in  However, when replacing only the linear layers of large networks, the model parameters count is significantly reduced, as demonstrated in our experimental results in Section~\ref{sec:cls}.

\textit{3) Communication Cost:}
The communication of \textit{Learn2pFed} includes two streams: a) the linear combination of model parameters, whose cost is same as that of most FL methods, and b) the loss value of each client, which can be neglected since it only costs 1 unit. Besides, the proposed method is scalable because both the communication cost and computational load increases linearly with the number of clients.

\section{Experiments} \label{sec:5}
In this section, we first conduct algorithm comparisons in a three-order polynomial regression task, and investigate the characteristics of $\textit{Learn2pFed}$ through ablation studies. Further, we apply it in both power consumption forecasting and image classification with the real-world data in various personalized FL settings and extend it to hundreds of clients, demonstrating superior performance compared to other baseline methods. All the experiments are implemented in PyTorch and simulated in NVIDIA GeForce RTX 3090 GPUs. Core codes are available in this link\footnote{https://github.com/kelenlv/Learn2pFed}.

\subsection{Experimental Setup}
\textbf{Baselines.}
We compare our proposed $\textit{Learn2pFed}$ with $12$ representative baselines under multiple experimental settings. Local-Only indicates that each client trains an independent model using its local data without federated collaboration. FedAvg~\cite{fedavg} and FedProx~\cite{fedprox} are two general FL baselines, while FedAvg+FT and FedProx+FT are their fine-tuning versions. Other personalized FL baselines include FedPer~\cite{fedper}, FedRep~\cite{fedrep}, Ditto~\cite{ditto}, pFedMe~\cite{pFedMe}, lp\_proj~\cite{lp-proj}, CFL~\cite{CFL}, and KNN-per~\cite{marfoq2022personalized}. Note that 
cluster-based personalized FL methods like CFL and KNN-per are only used in our classification tasks.

\textbf{Training Details.} We consider $500$ communication rounds of FL and $2$ epochs for each round with the batch size of $64$. We use Adam as the optimizer with a learning rate of $0.01$. For regression and forecasting tasks, we build up $\textit{Learn2pFed}$ following Alg.~\ref{alg:1} with $L=10$,  while using MLP and LSTM~\cite{yu2019review} as baseline models, respectively, for comparison. 
For image classification tasks, we use $\textit{Learn2pFed}$ as a plug-and-play model that replaces the last layer of the original CNN with a linear approximation; see more details in the supplemental materials.

\begin{figure*}[htbp]
  \begin{minipage}{0.58\linewidth}
    \captionof{table}{Regression performance w.r.t. three personalized FL settings on synthetic data. The proposed $\textit{Learn2pFed}$ achieves the best performance in all the three personalized FL settings.}\label{tab:simu}
    \centering
   \resizebox{\textwidth}{!}{%
   \begin{tabular}{l|c|ccc}
\toprule
Methods & Type in FL & \multicolumn{3}{c}{Averaged RMSE}\\
& & Setting 1 & Setting 2 & Setting 3\\
\midrule
Local-Only &- & 0.0204 & 0.0149 & 0.0208\\
\midrule
FedAvg & Generalized & $0.2067\pm 0.0070$ & $1.6571\pm0.1238$ & $3.9092\pm 3.8794$\\
FedProx & Generalized & $0.1351 \pm 0.0418$ & $0.4214 \pm 0.4953$ & $2.3072 \pm 1.5184$ \\
\midrule
FedAvg + FT & Finetune & $0.0023 \pm 0.0001$ & $0.0014 \pm 0.0001$ & $0.0716 \pm 0.0973$ \\
FedProx+ FT & Finetune & $0.0132 \pm 0.0179$ & $0.0176 \pm 0.0245$ & $0.0109 \pm 0.0150$ \\
FedPer & Split layers & $0.0006 \pm 0.0008$ & $0.0016 \pm 0.0007$ & $0.0029 \pm 0.0026$\\
FedRep & Split layers & $0.0175 \pm 0.0045$ & $0.0136 \pm 0.0023$ & $0.0154 \pm 0.0027$\\
pFedMe & Optimization & $0.0017 \pm 0.0002$ & $0.0111 \pm 0.0004$ & $0.0113 \pm 0.0008$\\
Ditto & Optimization & $0.0005 \pm 0.0000$ & $0.0011 \pm 0.0007$ & $0.0004 \pm 0.0000$ \\
lp\_proj & Optimization & $0.0023 \pm 0.0000$ & $0.0015 \pm 0.0000$ & $0.0017 \pm 0.0000$\\
\midrule
\textbf{\textit{Learn2pFed}} & Optimization & \bf{0.0002 $\pm$ 0.0002} & \bf{0.0003 $\pm$ 0.0002} & \bf{0.0003 $\pm$ 0.0002}\\
\bottomrule
\end{tabular}}
  \end{minipage}%
  \hspace{2mm}
  \begin{minipage}{0.38\linewidth}
    \centering
    \includegraphics[width=0.9\textwidth]{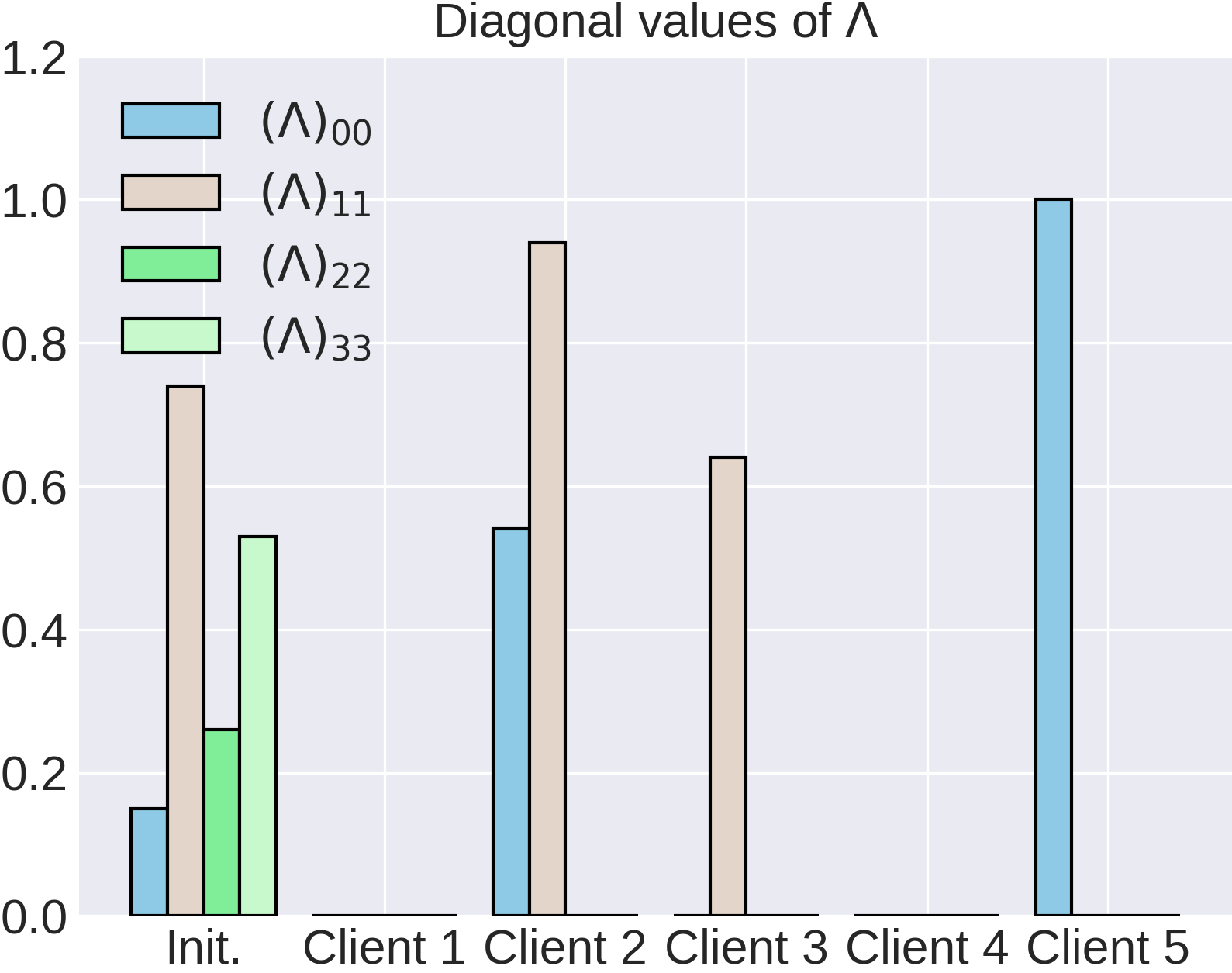} 
	\caption{Diagonal values of $\{\Lambda_i\}$. All clients share the same initialization of $\{\Lambda_i\}$ (on the left). The right shows the learned $\{\Lambda_i\}$ in five clients by $\textit{Learn2pFed}$ under Setting 1.}
	\label{fig:lambda}
  \end{minipage}
\end{figure*}

\subsection{Polynomial Regression Task}\label{sec:synthetic}
\textbf{Dataset and Federated Settings.} In this experiment, each client $i$ has a distinct ground-truth (gt) objective function $f_i(x)=\sum_{d=0}^3 \bm{a}_{i}[d]\cdot x^d$, where $\bm{a}_{i}=[a_0, a_1,a_2,a_3]$ is the polynomial coefficient vector. Different clients have different coefficient vectors, while they may share some coefficients. 
Here, we consider three different settings: 
\begin{itemize}
    \item Setting 1: all clients share three coefficients, i.e., $\bm{a}_i[d]=\bm{a}_j[d]; \forall i,j; \forall d \in \{0,1,2\}$.
    \item Setting 2: all clients share two coefficients, i.e., $\bm{a}_i[d]=\bm{a}_j[d]; \forall i,j; \forall d \in \{0,1\}$. 
    \item Setting 3: all clients share one coefficient, i.e., $\bm{a}_i[0]=\bm{a}_j[0]; \forall i,j$. 
\end{itemize}
The remaining coefficients are set distinctly across clients. Note that since high-order coefficients can have a greater impact on the disturbance of the function, we prefer to keep the lower-order coefficients the same across clients to increase the task's difficulty.
Finally, we generate local data by adding Gaussian noise to the local gt function with a mean of 0 and a standard deviation of 0.1.

\textbf{Results and Analysis.}  We perform the experiments for five independent trials with full 10-client participation, and report the averaged Root-Mean-Squared-Error~(RMSE) results in Table \ref{tab:simu}. 
We see that 1) the optimization-based methods, including Ditto~\cite{ditto} and lp\_proj~\cite{lp-proj}, perform better in terms of both accuracy and stability, as they exhibit smaller RMSEs and variances compared to other approaches.
However, the performance of the methods varies depending on the complexity of the dataset. For example, in simpler Setting 1, Ditto and FedPer show better performance than other methods, while in the more complex Setting 3, only Ditto outperforms other methods. This suggests that the choice of the method depends on the characteristics of the dataset.
2) Notably, $\textit{Learn2pFed}$ consistently outperforms other methods in fitting the polynomial model across all experimental settings, indicating that $\textit{Learn2pFed}$ is effective in capturing the underlying patterns in the personalized data and is robust to variations in the input.
More visualization results are shown in the supplemental materials.


\textbf{Impact of Learnable Parameters.}
We aim to investigate which specific learnable parameters play a more important role in the proposed $\textit{Learn2pFed}$ by repeating the simulations for five times and show the results in Table~\ref{tab:abla1}.
It reveals that learning more learnable parameters increases the representation power of $\textit{Learn2pFed}$ and improves the performance. 
Further, learning $\{\rho_i\}$ in~(\ref{eq:admm4}), which play a role like learning rates in forward propagation~(\ref{eq:layer1}), is shown to be not helpful enough.
However, learning the parameters $\{p_i, \Lambda_i\}$ (especially $\{\Lambda_i\}$) plays an important role in $\textit{Learn2pFed}$ since they are more concerned with the FL process. 
\begin{figure*}[ht]
  \begin{minipage}{0.6\linewidth}
    \captionof{table}{Learnable parameters ablation study w.r.t. three personalized FL settings on synthetic data: we leave the check marks under the learnable parameters and the blanks under the non-learnable parameters. The experiments with more learnable parameters perform better.}\label{tab:abla1}
    \centering
  \resizebox{\textwidth}{!}{
    \begin{tabular}{cccccc|ccc}
    \toprule
    \multicolumn{6}{c|}{Learnable parameters} &  \multicolumn{3}{c}{Averaged RMSE of $\textit{Learn2pFed}$}\\
    $\{\Lambda_i\}$&$\{p_i\}$&$\{\rho_i\}$&$\{\eta_i\}$&$\{\gamma_i\}$&$\{\theta_i\}$& Setting 1 & Setting 2 & Setting 3\\
    \midrule
    &&&&& &$0.0024\pm 0.0001$& $0.0023\pm 0.0001$& $0.0023\pm 0.0001$\\
    \checkmark&&&&& &$0.0007 \pm 0.0001$&$0.0010 \pm 0.0001$ &$0.0008 \pm 0.0002$ \\
    &\checkmark&&&&&$0.0018 \pm 0.0001$&$0.0226 \pm 0.0001$&$0.0100 \pm 0.0003$\\
    \checkmark&\checkmark&&&&& $0.0013 \pm 0.0001$ &$0.0008 \pm 0.0002$&$0.0058 \pm 0.0005$\\
     &&\checkmark&\checkmark&\checkmark&\checkmark&$0.0026 \pm 0.0024
    $&$0.0007 \pm 0.0003$&$0.0015 \pm 0.0003$\\
     &&\checkmark&&&&$0.0084 \pm 0.0006$&$0.0544 \pm 0.0037$&$0.0302 \pm 0.0154$\\
     &&&\checkmark&&&$0.0037 \pm 0.0006$&$0.0016 \pm 0.0004$&$0.0051 \pm 0.0000$\\
    &&&&\checkmark&&$0.0126 \pm 0.0009$&$0.0059 \pm 0.0013$&$0.0524 \pm 0.0005$\\
     &&&&&\checkmark&$0.0040 \pm 0.0004$&$0.0023 \pm 0.0002$&$0.0045 \pm 0.0021$\\
    &\checkmark&\checkmark&\checkmark&\checkmark&\checkmark&$0.0085 \pm 0.0071$&$0.0084 \pm 0.0038$&$0.0032 \pm 0.0019$\\
    \checkmark&&\checkmark&\checkmark&\checkmark&\checkmark&$0.0018 \pm 0.0001$&$0.0046 \pm 0.0034$&$0.0009 \pm 0.0003$\\
     \midrule
    \checkmark&\checkmark&\checkmark&\checkmark&\checkmark&\checkmark&$0.0002 \pm 0.0002$&$0.0003 \pm 0.0002$&$0.0003 \pm 0.0002$\\
    \bottomrule
    \end{tabular}}
  \end{minipage}%
  \hspace{2mm}
  \begin{minipage}{0.35\linewidth}
  \vspace{-5mm}
  \captionof{table}{Averaged RMSE for power consumption forecasting task. Lower is better. The proposed  $\textit{Learn2pFed}$ consistently performs the best under both personalized FL settings.}\label{tab:power}
    \centering
     \resizebox{\textwidth}{!}{
     \begin{tabular}{lcc}
        \toprule
        Methods & Setting 1& Setting 2\\
        \midrule
        Local-Only	&0.0998 $\pm$ 0.0002&$0.2166\pm 0.0001$\\
        FedAvg	&0.3796 $\pm$ 0.1535&$0.7465\pm0.2454$\\
        FedProx	&0.3799 $\pm$ 0.1533&$0.7471\pm0.2448$\\
        FedPer	&0.0342 $\pm$ 0.0177 &  0.1181 $\pm$ 0.0737 \\ 
        FedRep	&0.0341 $\pm$ 0.0178& 0.1182 $\pm$ 0.0737\\
        pFedMe	&0.0522 $\pm$ 0.0276&0.1226 $\pm$ 0.0413\\
        Ditto	& 0.0339 $\pm$ 0.0151& 0.1168 $\pm$ 0.0754\\
        lp\_proj& {0.0733 $\pm$ 0.0550}& {0.1577 $\pm$ 0.1200}\\
        \midrule
        \textbf{\textit{Learn2pFed}}	&\textbf{0.0307 $\pm$ 0.0001}&\textbf{0.0619 $\pm$ 0.0001}\\
        \bottomrule
        \end{tabular}}
  \end{minipage}
\end{figure*}

\textbf{Impact of $\{\Lambda_i\}$.}
We evaluate the impacts of learning $\{\Lambda_i\}$ by comparing the the performance of $\textit{Learn2pFed}$ achieved with and without learning $\{\Lambda_i\}$ in Setting 1 in Table~\ref{tab:simu}.
We find that learning $\{\Lambda_i\}$ greatly improves performance. Specifically, the averaged RMSE is decreased from 0.0026 to 0.0002, a $92\%$ reduction.
In addition, we analyze the learned $\{\Lambda_i\}$ in Fig~\ref{fig:lambda}. Notably, a 0 value of $\{\Lambda_i\}$ indicates the respective local parameters should adapt primarily to local data rather than actively participating in federated collaboration.
From Fig~\ref{fig:lambda}, we see that {the element of the matrix} $\{(\Lambda_i)_{33}\}\rightarrow0$ for all clients, which are consistent with our expectation since $\{(\Lambda_i)_{33}\}$ varies across clients in Setting 1 and thus should be learned locally.
Besides, it also suggests that $\bm{a}_i[2]$ (in the ground-truth objective function) need to be learned locally since $\{(\Lambda_i)_{22}\}\rightarrow0$. 
Overall, these findings highlight the importance of learning $\{\Lambda_i\}$ in our $\textit{Learn2pFed}$ algorithm and demonstrate its ability to adapt to the characteristics of the underlying local data distribution.

\textbf{Impact of the Number of Layers on Convergence.} 
Fig.~7 
in the supplemental materials shows the convergence of $\textit{Learn2pFed}$ in synthetic data w.r.t. three personalized settings mentioned above.
It demonstrates that the deeper network, which unrolls more iterations $L$ of the ADMM, leads to faster convergence and more accurate solutions.
Unless specified, we set $L=10$ for the subsequent experiments. 
\subsection{Power Consumption Forecasting}\label{sec:ELD}
\textbf{Dataset and Federated Settings.}  We use the dataset Electricity Consuming Load~\cite{ecl}~(ECL\footnote{https://archive.ics.uci.edu/ml/datasets/ElectricityLoadDiagrams20112014}) for electical load forecasting, which includes power consumption records (Kwh) for over 300 clients from 2011 to 2014.  
After data pre-processing, there are 313 candidate clients, each with 105216 records.
We perform experiments following two personalized FL settings: 
a) Setting 1 of full client participation scenario: we select $5$ clients that have the most distinct properties, which are distinguished by using t-SNE technique~\cite{van2008visualizing}.
b) Setting 2 of partial client participation scenario: we randomly sample $50$ clients of over 300 clients to participate at each FL round.
In both cases, we split the local data of each selected client into train and test subsets in a ratio of 9:1.

\textbf{Results and Analysis.} We conduct five independent trials and report the averaged RMSE results evaluated on the testing dataset in Table~\ref{tab:power}.
From the table, we see that different from the results in regression simulation task, optimization-based methods, including pFedMe~\cite{pFedMe} and lp\_proj~\cite{lp-proj}, fail to perform well in such real-world complicated datasets and require large tuning efforts. In contrast, $\textit{Learn2pFed}$ still outperforms the other approaches with lower RMSEs. Additionally, we provide visualizations of the prediction results for both participating and non-participating clients  
in the supplemental materials to verify the performance of the proposed $\textit{Learn2pFed}$.
\subsection{Image Classification}\label{sec:cls}
\textbf{Dataset and Federated Settings.}  We use two classical image classification datasets in FL, CIFAR-10~\cite{krizhevsky2009learning} and Fashion-MNIST~(FMNIST)~\cite{xiao2017fashion} in two personalized settings: 
\begin{itemize}
    \item 1) we consider full client participation with $M=10$ clients using the Dirichlet distribution~\cite{yurochkin2019bayesian} with argument $\beta_{dir}=\{0.1, 0.5\}$, where a smaller $\beta_{dir}$ indicates the greater heterogeneity among the clients.
    \item  2) we consider partial client participation in order to follow the convention in federated learning literature, e.g., in Ditto and FedRep. Specifically, we conduct experiments with 100 clients, and their data are generated according to the Dirichlet distribution with parameters $\beta_{dir}={0.1, 0.5}$. In each FL round, 10 clients are randomly selected for participation.
\end{itemize}
We further split the local data into training and testing sets at the ratio of 8:2 in both settings.
In order to leverage the powerful representation capabilities of the deep neural network, we use the features extracted from the second-to-last layer of a CNN as the input of $\textit{Learn2pFed}$. As a result, $\textit{Learn2pFed}$ aims to linearly estimate the last fully-connected layer. Then, we jointly train the CNN and $\textit{Learn2pFed}$ with only the latter involved in FL communication.

\begin{figure*}[ht]
  \begin{minipage}{0.67\linewidth}
    \captionof{table}{Averaged classification accuracy~(\%)  w.r.t. different Dirichlet parameters~($\beta_{dir}$) in CIFAR-10 and FMNIST in both settings, {and communication (Comm.) cost (KB) per epoch}. Our $\textit{Learn2pFed}$ consistently outperforms the state-of-the-art methods in both accuracy and Comm. cost, and the best results are in bold.}\label{tab:cls}
    \vspace{-2mm}
    \centering
   \resizebox{\textwidth}{!}{%
   \begin{tabular}{l|cccc|cccc||cc}
        \toprule
        Settings & \multicolumn{4}{c|}{10 clients} & \multicolumn{4}{c||}{100 clients} & \multicolumn{2}{c}{{{Comm. cost (KB) }}}\\
         \midrule
        Dataset & \multicolumn{2}{c}{CIFAR-10} & \multicolumn{2}{c|}{FMNIST} & \multicolumn{2}{c}{CIFAR-10} & \multicolumn{2}{c||}{FMNIST}&CIFAR-10&FMNIST\\
        \midrule
        $\beta_{dir}$ & 0.1&0.5& 0.1&0.5& 0.1&0.5& 0.1&0.5&-&-\\
        \midrule
        Local-Only&85.60&57.82&92.26&87.95 &71.25&50.43&92.20&87.46&0&0 \\
        FedAvg &30.05&31.01&76.04&77.87&30.69	&40.37	&	84.86&	83.24&62.01&10.29	\\
        FedProx&41.68&	52.54&	80.42&	86.19&52.56	&	48.22	&	90.84&	87.13&62.01&10.29	\\
        FedPer &89.12&	66.84 &96.55&91.67&84.08&		64.10&	97.54&	90.88&61.16	&5.28	\\
        FedRep & 86.56 & 62.39  &96.03&	88.72& 84.81	& 60.27& 	96.60 &		90.11&61.16&5.28	\\
        pFedMe&90.31&	65.19&	 97.48& 92.86&83.11	&	51.07&		98.15&	88.56&62.01&10.29	\\
        Ditto&87.30&	64.72&	 96.57&90.34&83.60&	54.87&	97.23&	89.24&62.01	&10.29	\\
        CFL&87.35&	64.29&96.89&	90.31&88.15&	51.90	&95.48&	89.60&62.01	&10.29\\
        kNN-Per   & 88.47 & 64.28   & 97.64  & 90.09 & 74.69& 	61.74	& 92.13& 	88.82&62.01&10.29	   \\
        \midrule
        \textbf{\textit{Learn2pFed}}& \textbf{90.71}&	\textbf{71.02}& \textbf{98.06}&\textbf{94.09}& \textbf{89.45}&	\textbf{71.64} & \textbf{98.97}&	\textbf{91.99}&\textbf{4.06}&\textbf{4.06}\\
        \bottomrule
        \end{tabular}}
  \end{minipage}%
  \hspace{2mm}
  \begin{minipage}{0.3\linewidth}
    \centering
    \includegraphics[width=1.0\linewidth]{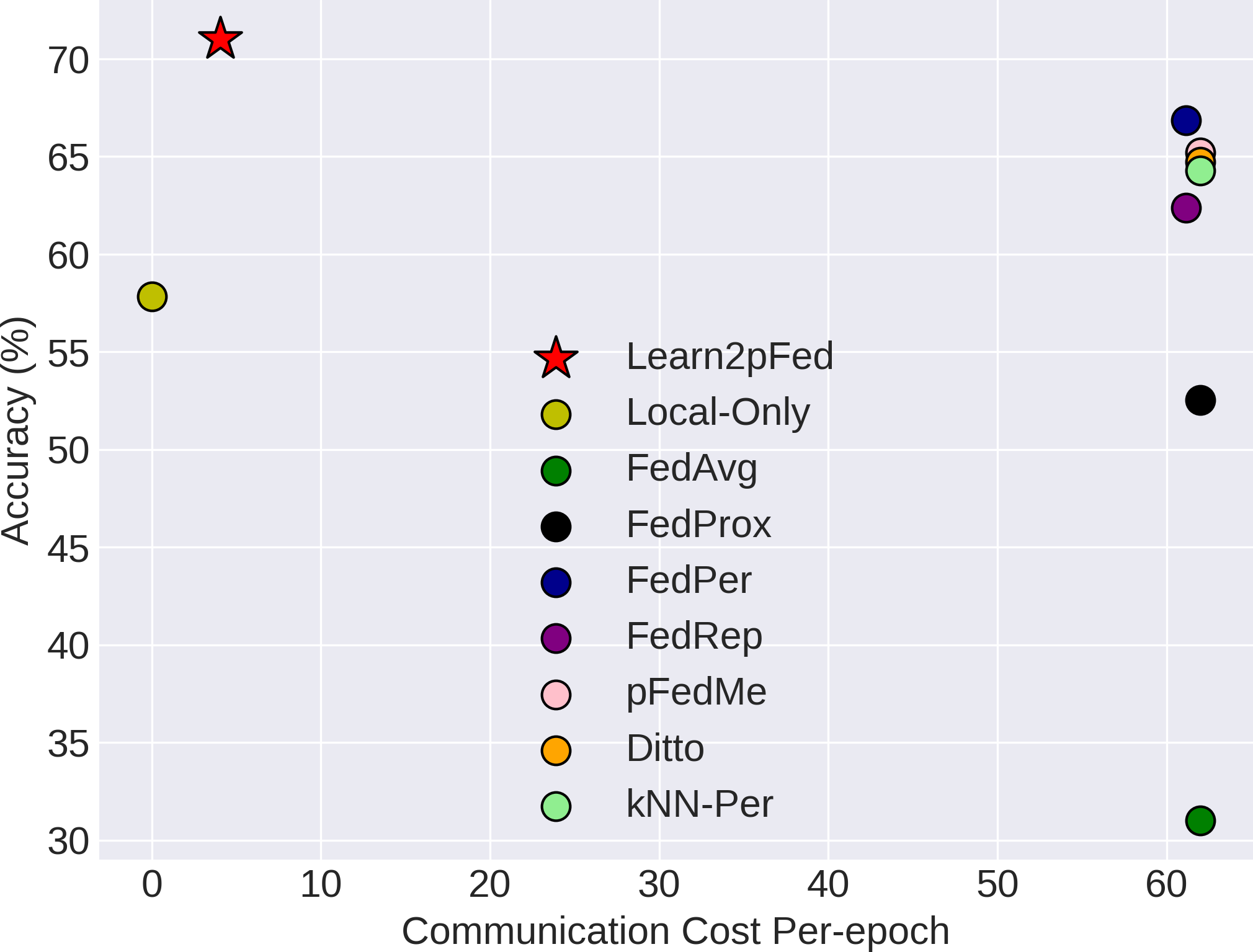}
  \caption{Communication cost and accuracy comparisons. \textit{Learn2pFed} achieves the highest accuracy with minor communication cost.}
  \label{fig:comm_acc}
  \end{minipage}
\end{figure*}

\textbf{Results and Analysis.} 1) Table~\ref{tab:cls} shows that our proposed $\textit{Learn2pFed}$ consistently outperforms baselines across different datasets and different levels of data heterogeneity, indicating the effectiveness of learning to determine which parts of parameters for federated collaboration. Besides, $\textit{Learn2pFed}$ demonstrates enhanced performance in the presence of increased data heterogeneity. Further experimental results are provided 
in the supplemental materials. 2) We observe that the performances of some previous optimization-based personalized FL methods drop significantly from Table~\ref{tab:cls} as the number of clients increases. The reason for this is about the intricate selection hyper-parameters, which in contrast, demonstrating the importance of dynamically determining the participation degree of local parameters in federated collaboration as our method does. 3) Fig.~\ref{fig:comm_acc} shows 
the communication cost per-epoch (KB) and accuracy of several representative methods in CIFAR-10 with 10 clients and $\beta_{dir}=0.5$. 
Combined with results in Table~\ref{tab:cls}, we see that our proposed $\textit{Learn2pFed}$ achieves the highest performance with minor communication cost, striking a great trade-off between communication cost and accuracy. This reveals another valuable property of our proposed $\textit{Learn2pFed}$ that it not only achieves pleasant accuracy, but also helps relieve communication cost. This reduction is attributed to $\textit{Learn2pFed}$ specifically replacing only the linear layers of the CNN model during federation, effectively minimizing communication overhead. Specifically, \textit{Learn2pFed} achieves 93.45\% reduction of communication cost, but outperforms FedAvg by 20\% in accuracy.

\begin{figure}[ht]
    \centering
    \includegraphics[width=0.4\textwidth]{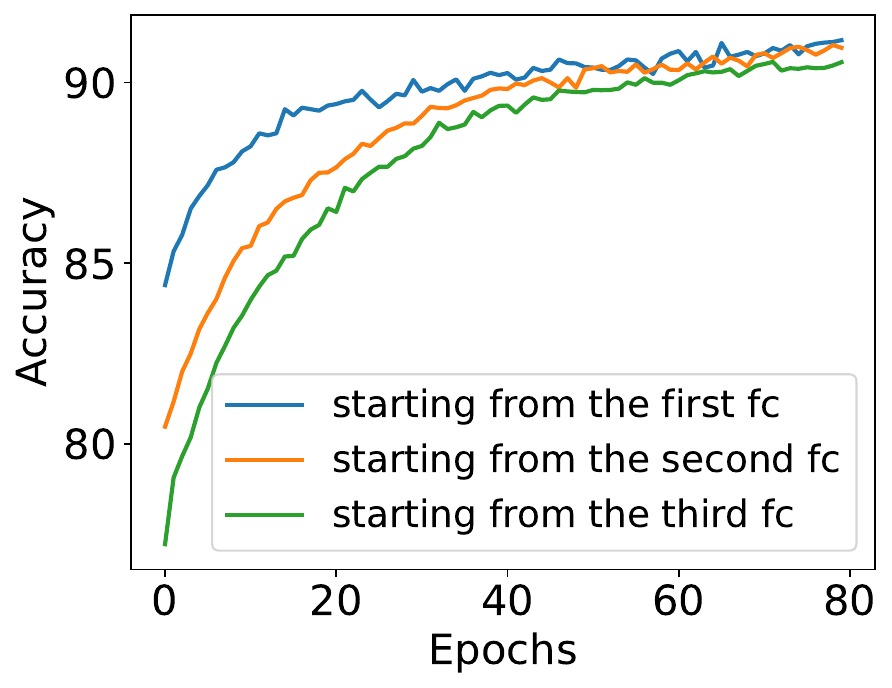}  
	\caption{Accuracy of $\textit{Learn2pFed}$ in CIFAR-10 with $\beta_{dir}=0.1$. Where $\textit{Learn2pFed}$ starts has little impacts on the accuracy in classification.} \label{fig:apd_cls}
\end{figure}

\textbf{Impact of layers where $\textit{Learn2pFed}$ starts in CIFAR-10.}
We conduct an ablation study on the selection of features extracted from the layer before the specific fc layers as inputs to $\textit{Learn2pFed}$ in CIFAR-10 ($M=10, \beta_{dir}=0.1$), i.e., where $\textit{Learn2pFed}$ starts. 
Table~\ref{tab:apd_cls} shows the quantitative results. 
When $\textit{Learn2pFed}$ starts from the first fully-connected layer, the shared ratio increases significantly because the basic CNN's parameter quantity primarily concentrates on the first fc layer. 
For this reason, our method may not always demonstrate a significant ability to reduce communication costs when starting at different layers. However, it does not impact much on the accuracy~(see in Fig.~\ref{fig:apd_cls}). 
\begin{table}[ht]
    \caption{Ablation study on layers where $\textit{Learn2pFed}$ starts. The shared ratio denotes the ratio of local parameters to shared parameters where the latter is fixed in $\textit{Learn2pFed}$. The fully-connected layers' input and output dimension~($dim$) follows the architecture introduced 
    in the supplemental materials.}\label{tab:apd_cls}
    \centering
   \begin{tabular}{ccc|cc}
        \toprule
        \multicolumn{3}{c|}{fully-connected layers} & local  & shared \\
        first $dim$ &second $dim$&third $dim$& parameters & ratio\\
        {[400,120]}&[120,84]&[84,10]& (KB)  & (\%)\\
        \midrule
        &&\checkmark & 61.14&14.72\\
        &\checkmark& &50.99& 17.65\\
        \checkmark&&&2.87&313.58\\
        \bottomrule
        \end{tabular}
\end{table}

\textbf{Discussions on Resource Usage.} We present the memory and computation costs in CIFAR10 classification in Table~\ref{tab:param}, comparing them to those of FedAvg. Due to the replacement of the last few linear layers of the network with \textit{Learn2pFed} in the classification task,  we significantly cut memory usage by 88.82\% (with feature extractor) and 93.13\% (without feature extractor). Additionally, there is a 9.04\% reduction in floating-point operations (FLOPs).
That is, despite the reduction in resource usage, our performance has shown a notable improvement, demonstrating the effectiveness of our approach.

\begin{table}[ht]
    \caption{The storage burden of local parameters.}
    \label{tab:param}
    \centering
    \begin{tabular}{c|cc|c}
    \toprule
         &\multicolumn{2}{c|}{Memory (KB)}& \multirow{2}{*}{FLOPs (M)}\\
         &  Feature Extractor & Linear layers  \\
         \midrule
         FedAvg & 2.87 &59.13&0.6517\\
     \textit{Learn2pFed}&2.87&\textbf{4.06}&0.5928\\
         \bottomrule
    \end{tabular}
\end{table}
\vspace{-1em}

\section{Conclusion}\label{sec:con}
We introduce $\textit{Learn2pFed}$, a novel framework for personalized federated learning through algorithm unrolling. 
Our framework tackles the challenge of learning hyper-parameters that are typically unlearnable in the optimization process. 
By allowing the learnable parameters to determine the participation of local models in federated learning, we enhance  
adaptability of personalized FL methods. Extensive experiments on synthetic, time-series, and natural image datasets demonstrate the superior performance of $\textit{Learn2pFed}$. Furthermore, as the unrolling-based framework, it holds potential for application in various scenarios in personalized FL approaches.

$\textit{Learn2pFed}$ focuses on dynamically determining the local parameters that should participate in the federated collaboration, but a limitation arises in its ability to explain the physical meaning of those parameters selected by$\textit{Learn2pFed}$. We are intrigued by the potential insights it may offer for model compression or data selection and aim to explore this further in future work.


\newpage

\bibliographystyle{IEEEtran}
\bibliography{refs}


 




\vfill

\end{document}